\documentclass{vgtc}
\ifpdf
  \pdfoutput=1\relax                   
  \usepackage{graphicx}                
  \DeclareGraphicsExtensions{.pdf,.png,.jpg,.jpeg} 
\else
  \usepackage{graphicx}                
  \DeclareGraphicsExtensions{.eps}     
\fi%

\graphicspath{{figures/}{pictures/}{images/}{./}} 

\usepackage{microtype}                 
\PassOptionsToPackage{warn}{textcomp}  
\usepackage{ulem}
\usepackage{comment}
\usepackage[explicit]{titlesec}
\usepackage{adjustbox}
\usepackage{textcomp}                  
\usepackage{mathptmx}                  
\usepackage{times}                     
\usepackage{cite}                      
\usepackage{tabu}                      
\usepackage{booktabs}                  

\usepackage{wrapfig}
\usepackage{subfig}
\usepackage{tabularx}
\usepackage{graphicx}
\usepackage{xcolor}
\usepackage{enumitem}

\usepackage{flushend}

\newcommand{\new}[1]{\textcolor{black}{#1}}

\onlineid{0}

\vgtccategory{Research}

\vgtcinsertpkg

\title{Bio-inspired Structure Identification in Language Embeddings}

\author{Hongwei (Henry) Zhou\thanks{e-mail: hzhou55@ucsc.edu}
\qquad  Oskar Elek\thanks{e-mail: oelek@ucsc.edu, website: https://cgg.mff.cuni.cz/\texttildelow oskar/}
\qquad  Pranav Anand\thanks{e-mail: panand@ucsc.edu, website: https://people.ucsc.edu/\texttildelow panand/}
\qquad  Angus G. Forbes\thanks{e-mail: angus@ucsc.edu, website: https://creativecoding.soe.ucsc.edu/}
\\{\scriptsize University of California, Santa Cruz}%
}

\teaser{
  \centering
  \vspace{-5mm} 
  \includegraphics[width=\columnwidth]{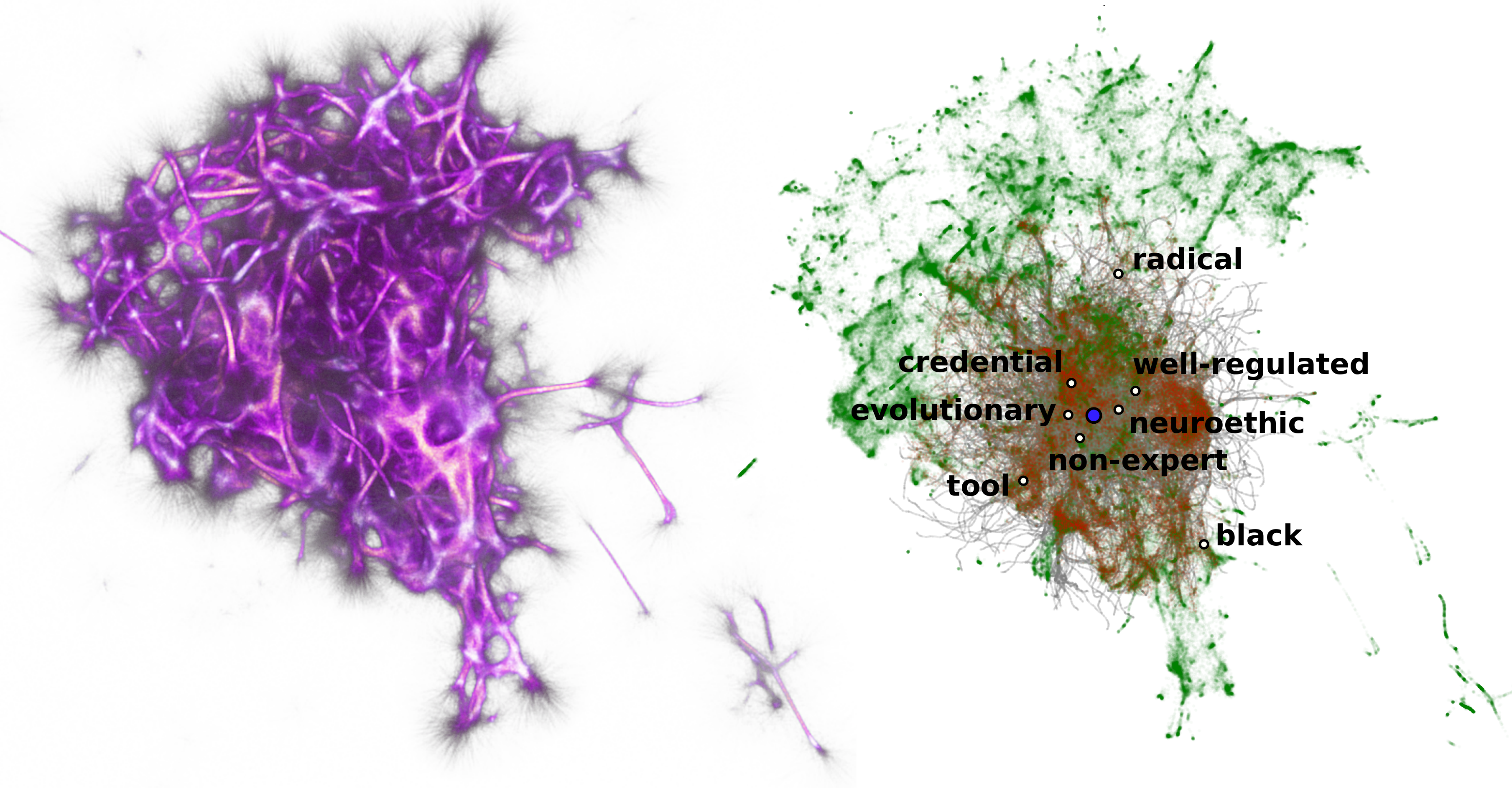}
  \label{fig:teaser}
  \vspace{-8mm}
  \caption{Left: 3D volume color-map visualization of the geometric structures recovered from Gensim Continuous Skipgram embedding of Wikipedia 2017 (300k tokens). Right: structure-guided exploration of the intermediate neighborhood of the noun \textit{research}, highlighting the discovered tokens (red) from among all tokens (green). \new{Tokens are discovered by structure-guided agents that stochastically explore the embedding space.}}
}

\abstract{
Word embeddings are a popular way to improve downstream performances in contemporary language modeling. However, the underlying geometric structure of the embedding space is not well understood. We present a series of explorations using bio-inspired methodology to traverse and visualize word embeddings, demonstrating evidence of discernible structure. Moreover, our model also produces word similarity rankings that are plausible yet very different from common similarity metrics, mainly cosine similarity and Euclidean distance. We show that our bio-inspired model can be used to investigate how different word embedding techniques result in different semantic outputs, which can emphasize or obscure particular interpretations in textual data.
} 

\CCScatlist{
  \CCScat{}{Human-centered computing}{Visualization}{Visualization techniques}{};
  \CCScat{}{Computing methodologies}{Artificial intelligence}{Natural language processing}{};
}

\titleformat{name=\paragraph,numberless}[runin]
  {\normalfont\normalsize\bfseries}{}{1pt}{\uline{#1.}}

\begin{document}

\firstsection{Mapping Language}

\maketitle

Much work extracting meaning from text has relied on relational structures that can be represented (and visualized) as graphs. Phrase Nets~\cite{vanHam2009phrasenets}, for instance, uses nodes to represent words (`tokens') and edges for the user-defined relations between them. Depending on the interpretation of the working data, higher-level entities can be mapped to this structure, such as documents~\cite{PerezMessina2018documentevolution}, stories~\cite{Subasic2008graphstories}, or even ideas~\cite{Onduygu2020philosophy} -- with suitable relational axioms applied to them. At a more granular level, syntactic relations in linguistics are often represented as graph diagrams~\cite{Partee1993MathMethods}, as are the ontological relationships between words~\cite{Fellbaum1998wordnet}. While such relational structures have proven incredibly valuable, they are difficult to automatically generate from text, a problem since there are often countless relations one might wish to extract from a text. Moreover, from a computational perspective, graphs can be very costly: even in the simple example of the Bigram model, the resulting relationship graph will have $O(N^2)$ edges in a dataset with $N$ tokens. For more complex relations, this cost can grow still further, rendering graphs difficult to handle for datasets with $10^4$--$10^5$ or more tokens.

\begin{figure*}[t]
\begin{center}
\includegraphics[width=2.1\columnwidth]{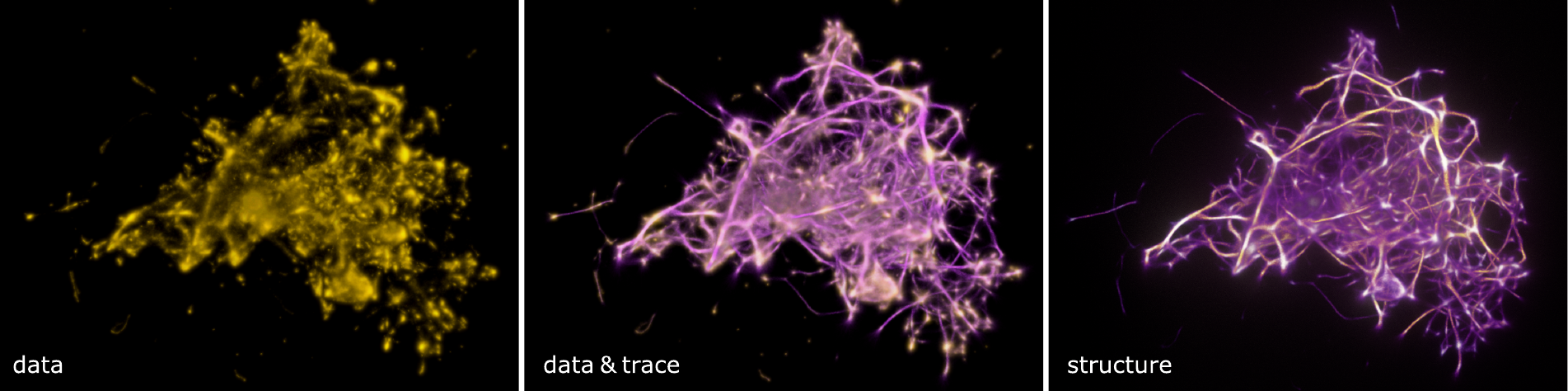}
\vspace{-6mm}
\caption{3D volume visualization of global embedding for W2V-300k. Left: The token data are represented as a density field, rendered in yellow. Middle: geometry discovered in these data by MCPM is represented by another density field -- trace -- overlaid over the data in purple. Right: the structure of the trace field is rendered as an emissive 'heatmap', with the highest density values also emitting more light.}
\label{fig:global_embedding_3D}
\vspace{-6mm}
\end{center}
\end{figure*}

In recent years, word embeddings -- such as Word2Vec, GloVe, ELMo, and BERT -- have gained real remarkable traction as representations of word-level information. Their key computational idea is to transform topological information contained in a relational graph to geometric information encoded in a D-dimensional vector (`embedding’) space by using a deep learning model. This brings the representational cost down from $O(N^2)$ to $O(ND)$, where $D \ll N$, typically in the 100s. On top of the efficiency increase, embeddings have a number of interesting algebraic properties: most importantly, the contextual similarity of the embedded tokens is transformed into geometric proximity in the embedding \cite{mikolov2013distributed, coenen2019visualizing}. Because they explicitly consider the token's context~\cite{devlin2018bert, radford2019language}, it has been shown that embeddings contain information that can be processed to extract a range useful properties: clustering by token usage~\cite{rogers2020primer, wiedemann2019does} as well as different kinds of syntactic information~\cite{rogers2020primer, lin2019open, hewitt2019structural}. Thus, there is the promise that this kind of method could provide high-dimensional representations that encode a large manner of relations implicitly without having to hand-code them in advance.

In spite of the progress in language processing tasks, understanding the information contained in embeddings is still challenging due to their high dimensionality. While parallel coordinates are suitable for high dimensional data \cite{Dou2011parallelprobabilistic, Collins2009parallelcourts} they do not capture the spatial relationships critical in embeddings. Therefore, the standard way to visualize embeddings is currently to project the token data to 2D or 3D using PCA, UMAP and other dimensionality reduction techniques \cite{EmbeddingProjector}, optionally with additional semantic annotations \cite{coenen2019visualizing}. In that process, two different distortions happen to the data: distortion of high-level structure, and induction of relations that are not part of the original embedding. The inclusion of explicit referencing information between the tokens \cite{Berger2016cite2vec} and identification of salient dimensions \cite{Ji2019embeddingexploration} does seem to leverage some of these issues. \new{In addition, interactive visualization tools have been proposed for literary experts and natural language processing researchers~\cite{liu2019latent, heimerl2018interactive}. These focus on exploring linear relationship between word embeddings, identifying concepts and experimenting with attribute vectors.}

Yet our understanding of the embeddings’ structural properties remains far from complete \cite{Kucher2015textvisbrowser}. For instance, recent work on contextual embeddings has found that the embedded tokens have a highly anisotropic distribution, which impacts the standard cosine distance used to measure their closeness \cite{wiedemann2019does}. At least in the case of non-contextual (`global’) embeddings, normalizing the distribution to be more isotropic and centered about the origin improves the performance of downstream tasks that build on this way of measuring distances \cite{mu2017all}. These and other results show that there are gaps in our understanding of the various embedding data.


\section{Language as organic structure}

We propose a visualization and analysis framework for language embeddings based on bio-inspired \textbf{optimal transport networks}. The mathematics of optimal transport \cite{Villani2009optimal, peyre2019computational} is based on the principle of least effort, which applies to phenomena ranging from particle and light transport to the behavior of living beings. Ubiquitous in nature, we posit that this principle applies to language alike -- an idea already explored by Zipf~\cite{Zipf1949leasteffort} and others since.

We make the following key assumption: that the relational structure of language is reasonably preserved in (or transferred into) the embedded representations. Building on that, we look into the possibility of discovering the geometric structures in both local and global embeddings (Section \ref{sec:dataset}). Our contributions towards this include:
\begin{itemize}
	\item Recovering geometric structures in these data by using a combination of dimensionality reduction techniques (mainly UMAP) and a pattern-finding algorithm based on optimal transport in biological systems dubbed MCPM (Figure~1-left, Section \ref{sec:structure}).
	\item Design of a custom random-walk exploration technique to examine the recovered structures through the lens of a few standard language processing tasks (Figure~1-right, Section \ref{sec:exploring}).
	\item Demonstration of the utility of these techniques for both gaining insight into the embedding data and improving the performance of standard language processing tasks (Section \ref{sec:discussion}).
\end{itemize}
Ultimately, this work is a first step towards developing a framework to enable human-readable exploration of machine-generated language data. Section \ref{sec:future} discusses the future steps we plan to undertake to make this happen.


\begin{figure*}[t]
\begin{center}
\includegraphics[width=2.1\columnwidth]{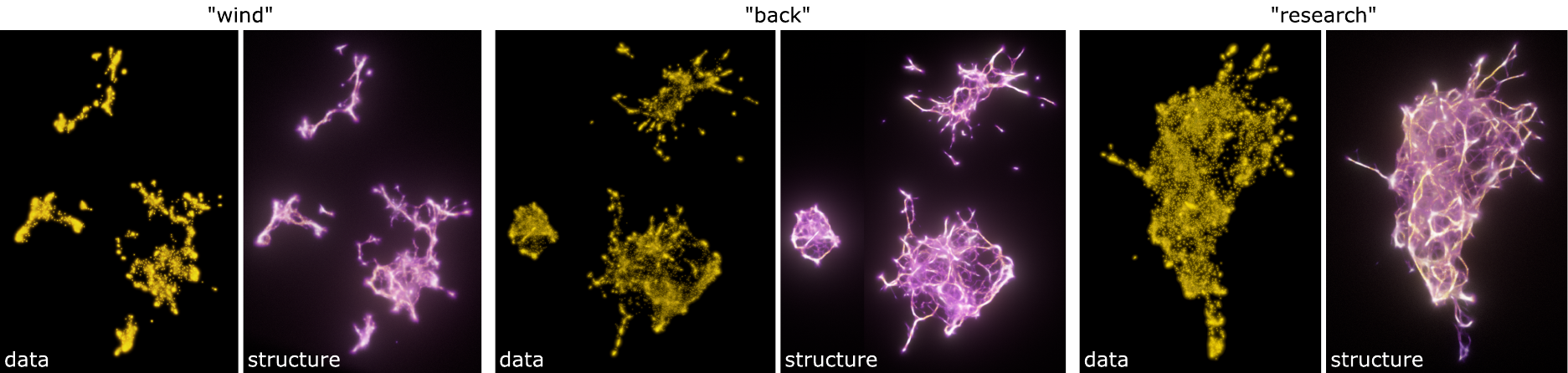}
\vspace{-6mm}
\caption{3D volume visualization of local embeddings for words \textit{wind}, \textit{back} and \textit{research}. The modalities are identical with \protect{Figure~\ref{fig:global_embedding_3D}} (with the joint visualization of data and trace skipped). \new{For each token, we observe different numbers of clusters and their shape.}}
\label{fig:local_embedding_3D}
\vspace{-6mm}
\end{center}
\end{figure*}

\section{Datasets} \label{sec:dataset}

Language embedding data can be generated through a variety of embedding algorithms applied to specific text corpora. To take the examples of BERT \cite{vaswani2017attention} and Word2Vec \cite{mikolov2013distributed}, both propose novel neural network architectures to vectorize English words. The networks are both trained to guess masked words given its surrounding word tokens. The assumption behind these systems is that in the embedded space each word’s surroundings can capture its context.

We select these two language embedding models -- BERT (Figure~\ref{fig:local_embedding_3D}) and Word2Vec (Figure~\ref{fig:global_embedding_3D}) -- and generate the following datasets that are the basis for our further analysis.

\paragraph*{Local (contextual) embeddings } Similar to Coenen et al., this dataset uses base BERT for embedding generation and Wikipedia as language corpus \cite{coenen2019visualizing}. Each generated dataset is particular to a single word, and defines the context relative to that word -- typically resulting in 1000s of tokens. Each data point then refers to a sentence in which the word occurred within the corpus. We can interpret this as a volume of semantic space where the meaning of a token can fluctuate based on its actual usage in the text. We will be referring to these as “BERT-X”, with “X” being the represented word.

\paragraph*{Global embeddings} This dataset uses Gensim Continuous Skipgram -- a variation of Word2Vec -- to process the English Wikipedia Dump of February 2017, resulting in approximately 300k tokens. In contrast to contextual embeddings, each data point refers to a single token instead of the sentence in which a token is used. A token includes two pieces of information: the word and its part of speech. For example, \textit{wind\_NOUN} and \textit{wind\_VERB} are considered separate tokens and occupy different positions. This is different from BERT where the part of speech is not made explicit. We will be referring to this dataset as ``W2V-300k''.

\paragraph*{Dimensionality reduction} The original word embeddings are high-dimensional: BERT embeddings are 768-dimensional while those by Gensim Continuous Skipgram are 300-dimensional. To make visualization and analysis feasible, we rely on UMAP (with neighborhood size of 15) to project the data to 3D space. \new{The dimensionality reduction is necessary, due to the high memory requirements of our notion of transport network: this is based on a continuous representation by a \textbf{density field}, both in the visualization (Section~\ref{sec:structure}) and exploration (Section~\ref{sec:exploring}) stages.}

Naturally, this implies the loss of some information, particularly sacrificing the global structure of the data to preserve local neighborhood relations (however, less so than other dimensionality reduction methods such as PCA or TriMap \cite{Amid2019trimap}). It is also likely that additional geometric structures are induced in this process. Yet in Section~\ref{sec:exploring} we show that even these distortions do not render the dataset unusable when the detected structures are used for navigating the embedding. \new{Sections \ref{sec:discussion} and \ref{sec:future} discuss this in further detail.}


\begin{figure*}[tb]
\begin{center}
\includegraphics[width=2\columnwidth]{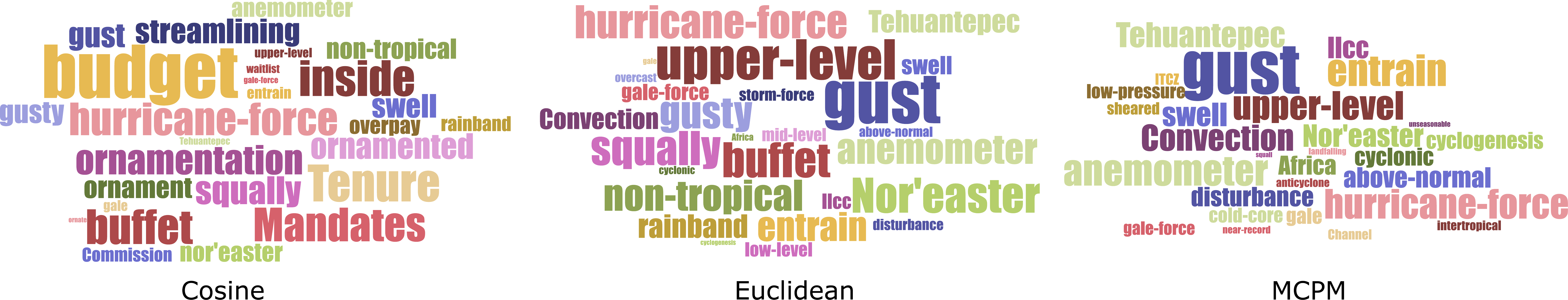}
\vspace{-2mm}
\caption{Word clouds of top 30 most similar words for \textit{wind\_NOUN} in the W2V-300k according to three similarity metrics. The bigger the word, the more similar it is to \textit{wind\_NOUN}.}
\label{fig:wordcloud}
\vspace{-6mm}
\end{center}
\end{figure*}

\section{Structure Detection and Visualization} \label{sec:structure}

Having the embedding data projected to 3D as a point cloud (Section \ref{sec:dataset}), the next step is to find a transport network that spans it. For this purpose we use the Monte Carlo Physarum machine (MCPM), a \textbf{pattern-finding algorithm} inspired by the growth and foraging behavior of Physarum polycephalum `slime mold’ \cite{elek2020monte}. This method has been previously applied in astronomy for inferring the quasi-fractal structure of the cosmic web \cite{burchett2020revealing, simha2020disentangling}, where it has successfully recovered the theoretically predicted filamentary patterns over sparse galaxy data.

MCPM is a hybrid method, in which a swarm ($10^6$--$10^7$) of discrete agents explores a domain represented by a continuous 3D lattice. This lattice stores the spatial footprint of the input data, which then acts as an attractor for the agents. As a result, the agents interconnect the input data in a single continuous transport network. This emergent network is represented by another lattice referred to as \textbf{trace}, effectively storing the scalar spatio-temporal density of the model’s agents. This representation of the transport network is advantageous for our further analysis (Section \ref{sec:exploring}), serving as a guidance mechanism for exploring the connections between different embedding tokens or, generally, distinct regions in the embedding.

\paragraph*{Visualization} To visualize the overall 3D network structure, we rely on a combination of direct volume rendering \cite{RezkSalama2001volume} and physically-based volumetric path tracing \cite{Pharr2016pbrt}. \new{The main tasks the visualization addresses at this stage are:
\begin{itemize}[leftmargin=5mm,nosep]\setlength\itemsep{0em}
    \item understanding the overall structure of each embedding,
    \item identifying the number of distinct and significant clusters in the analyzed embedding, as well as their shape, and
    \item recognizing whether different embeddings contain similar structural patterns and on what scales they are present.
\end{itemize}}
\noindent In the remainder of this section, we focus on a qualitative analysis of the global and local embedding datasets introduced in Section \ref{sec:dataset}, \new{through the lens of the above tasks}. Section \ref{sec:exploring} then focuses on a finer, \new{token-level} exploration of these data.

\paragraph*{Global embeddings} Even though the token data in W2V-300k appear disorganized on the first glance (Figure~\ref{fig:global_embedding_3D}), MCPM reveals rich network-like geometry. This structure is chaotic, even fractal, with filaments folded into themselves -- probably as a result of compressing the high-dimensional embedding data into 3D. The structures exist at the level of entire clusters, rather than interconnecting individual words, although this might be a limitation of the underlying lattice resolution.

Even though some outlying tokens are present, the vast majority of these data forms a single densely interconnected network. This might be a reflection of how interlinked the source Wikipedia corpus is. We observe two recurrent features: \textbf{filaments} and \textbf{knots}. Some token clusters are distributed along the filaments, while others lie in the knots where multiple filaments intersect. We also notice different strengths (densities) of filaments, usually in proportion to the number of tokens contained within them.

\begin{figure}[b]
\begin{center}
\includegraphics[width=0.8\columnwidth]{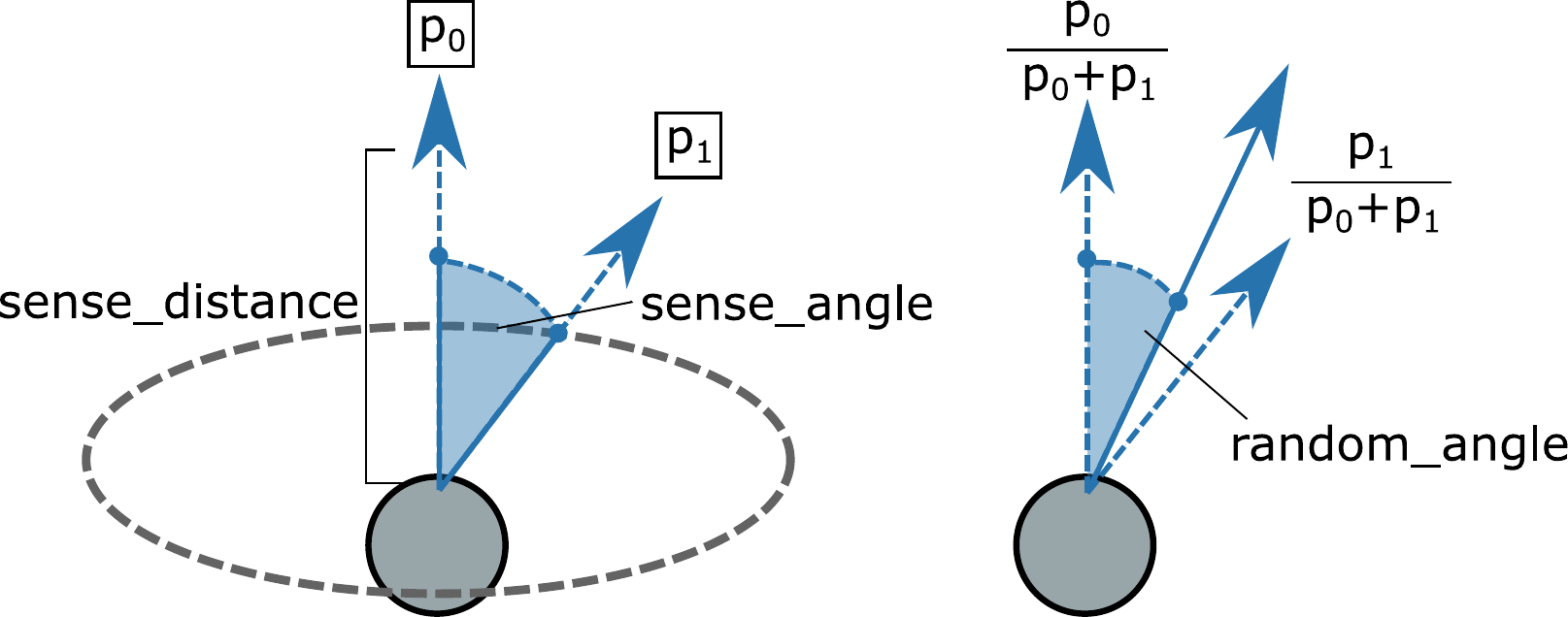}
\vspace{-2mm}
\caption{Illustration of the probe agent behavior: sensing phase (left) and steering phase (right). The values $p_0$ and $p_1$ are sampled from the trace field.}
\label{fig:agentbehave}
\vspace{-6mm}
\end{center}
\end{figure}

\begin{figure}[tb]
\begin{center}
\includegraphics[width=\columnwidth]{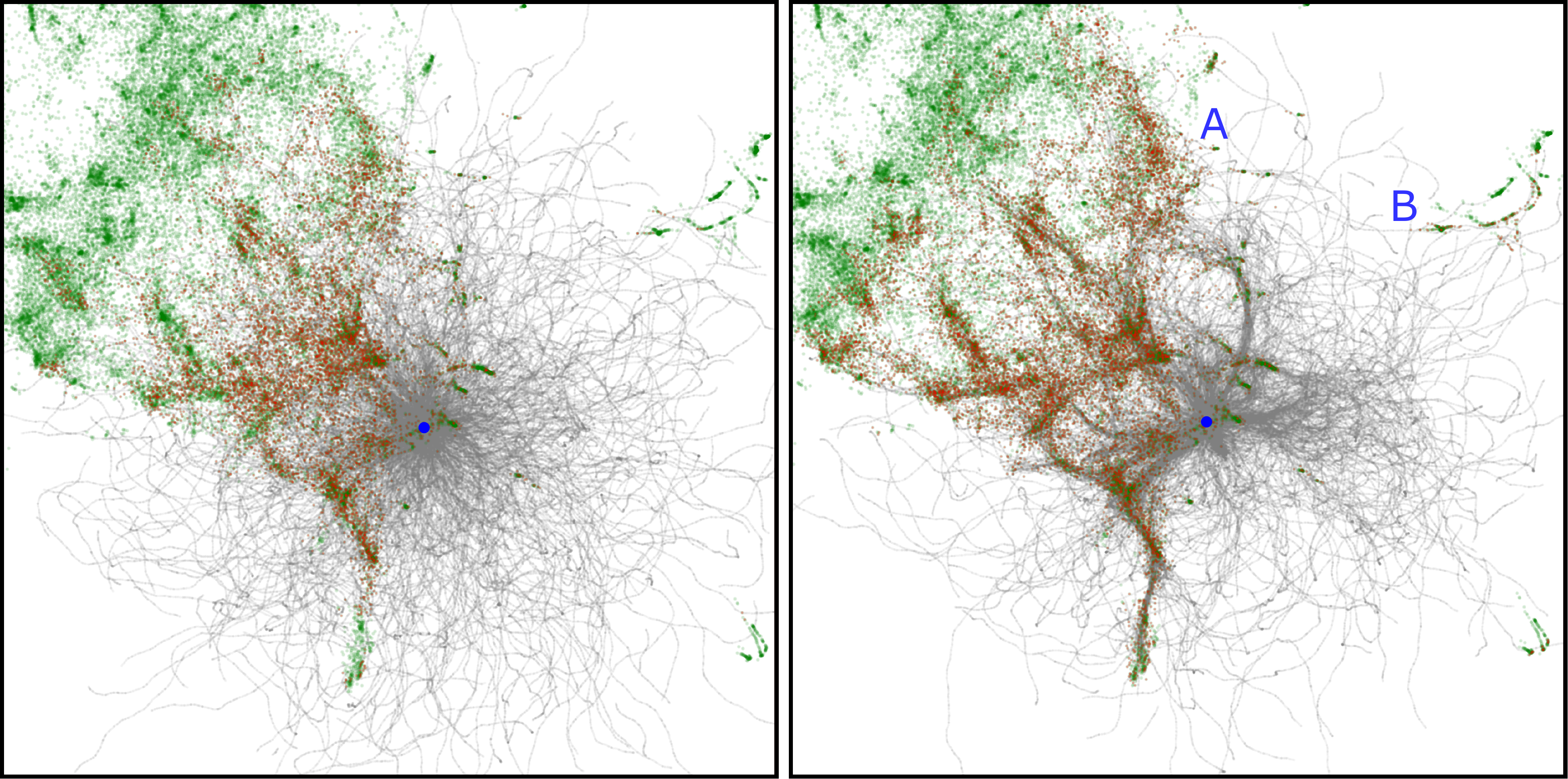}
\vspace{-6mm}
\caption{MCPM agent exploration results for token \textit{class\_NOUN} in W2V-300k comparing unguided (left) and trace-guided traversal (right). The agents are spawned at the same starting position (blue dot) and their trajectories are marked in grey. They are set out to discover the green data points (tokens), which are marked in red when discovered. To avoid cluttering we only draw a subset of the token data (within a narrow slice centered around the starting point), but still draw all the agent trajectories to emphasize the patterns of their movement.}
\label{fig:tracecompare}
\vspace{-6mm}
\end{center}
\end{figure}

\paragraph*{Local embeddings} Since each local embedding corresponds to a specific word, we have generated the BERT embeddings for \new{several interesting words and then chose} three representative ones: \textit{wind} (a homonym), \textit{back} (a polysemous word), and \textit{research}. The geometric structures that MCPM infers here (Figure~\ref{fig:local_embedding_3D}) are similar to W2V-300k, but exist on a smaller scale: the filaments interconnect words or groups of words, rather than entire clusters.

MCPM also functions as \new{an implicit} \textbf{clustering mechanism}. UMAP is designed to preserve components during the dimensionality reduction, but clustering them is not trivial due to their irregular shapes. MCPM manages to identify and interconnect the clusters (as a function of the model’s characteristic feature scale). Some clusters are interconnected densely similar to W2V-300k, others are sparse and branchy, perhaps indicative of the usage patterns of these words. The number of connected components also matches intuition: words like \textit{wind} and \textit{back} have multiple discrete contexts, while \textit{research} has only one but very broad context. General terms like \textit{class} and \textit{suit} yield more than 10 components. \new{Further discussion on clustering is provided in Section~\ref{sec:discussion}.}


\section{Exploring the Embedding}\label{sec:exploring}

Having extracted the trace field representing the \textbf{transport network} over the embedded tokens, this section covers the mechanisms of navigating this structure and the resulting word similarity measure we propose based on it. For this purpose we adopt the following metaphor: we interpret the transport network as an electrically conductive medium. Filaments with high density offer high throughput, while low-density areas (lacunae) have high resistance. The shortest path in such a structure is one that minimizes travelling distance and maximizes throughput.

\paragraph*{Navigating the trace} We deploy an agent-based algorithm inspired by MCPM (see Section \ref{sec:structure}), but significantly simplified. We will refer to the agents of this process as \textbf{probe agents}. The main difference is that probe agents traverse the already detected trace field without modifying it. Second, their geometric behavior is more basic in comparison to MCPM agents.

Each step of the probe agents consists of two phases: \textbf{sensing} and \textbf{steering}. In the sensing step, an agent samples values $p_0$ and $p_1$ from the trace (Figure~\ref{fig:agentbehave}-left). The sample distance $sense\_distance$ is determined prior to the simulation. The value $p_0$ lies along the agent’s current movement direction, while $p_1$ is sampled from a cone determined by a constant $sense\_angle$. Then in the steering step, the agent makes a decision whether to turn or not based on the probability proportional to $p_0$ and $p_1$ (Figure~\ref{fig:agentbehave}-right). If the agent turns, its new movement direction is then \new{changed} by $0 < random\_angle < sense\_angle$ towards the sensing direction, \new{with $random\_angle$ sampled uniformly in the given interval.}

Each probe agent’s behavior is a \textbf{random walk process}. Due to the probabilistic steering step, the trace guides the agents so that they effectively follow the transport network structure. Our typical random walk search uses 900 probe agents, each performing 500 steps. We consider a token `discovered’ if any of the agents passes around it within a small distance, usually between $1/400$ and $1/200$ of the domain size.

\begin{table}[tb]
\centering
\caption{Ranking difference list between MCPM, Euclidean and cosine rankings. The entries are ordered by difference of MCPM and Euclidean rankings.}
\label{tb:diff}
\vspace{-2mm}
\begin{tabularx}{\columnwidth}{lccc}
\textbf{Word} & \textbf{MCPM Rank} & \textbf{Euclid Rank} & \textbf{Cosine Rank}
\tabularnewline\hline
unseasonable & 28 & 271 & 908 \tabularnewline\hline
near-record & 26 & 262 & 796 \tabularnewline\hline
anticyclone & 25 & 65 & 181 \tabularnewline\hline
intertropical & 24 & 44 & 125\tabularnewline\hline
squall & 29 & 49 & 138 \tabularnewline
\end{tabularx}
\end{table}

\begin{table}[tb]
\caption{Three samples from each of the three major clusters detected in BERT-back. See \protect{Figure~\ref{fig:localandexplore}} for the corresponding visualization.}
\label{tb:cluster}
\vspace{-6mm}
\begin{tabularx}{\columnwidth}{p{25.5em}}
\tabularnewline
\multicolumn{1}{c}{\textbf{Top cluster}} \tabularnewline 
\hline
Partition walls constructed from fibre cement \textit{backer} board are popular as bases for tiling in kitchens or in wet areas like bathrooms. \tabularnewline \hline
At one time a firm called Submarine Products sold a sport air scuba set with three manifolded \textit{back-mounted }cylinders.\tabularnewline \hline
The harnesses of many diving rebreathers made by Siebe Gorman included a large \textit{back-sheet} of reinforced rubber.
\vspace{2mm}
\tabularnewline
\multicolumn{1}{c}{\textbf{Bottom-left cluster}} \tabularnewline
\hline
Mono Lake is believed to have formed at least 760,000 years ago, dating \textit{back} to the Long Valley eruption. \tabularnewline \hline
Other settlements were Toro, in the extreme south, 1827, and Noble, in the north portion, dating \textit{back} to the 1830s. \tabularnewline \hline
Early history: The area comprising the city of Bell has a Native American history dating \textit{back} thousands of years.
\vspace{2mm}
\tabularnewline
\multicolumn{1}{c}{\textbf{Bottom-right cluster}} \tabularnewline 
\hline
Decisions must be unanimous: any divided decision sends the question \textit{back} to the House at large. \tabularnewline \hline
He ends by saying that, if he does not hear \textit{back} from Romani, he will not write to him again.\tabularnewline \hline
Cartoons often would be rejected or sent \textit{back} to artists with requested amendments, while others would be accepted and captions written for them.\tabularnewline
\end{tabularx}
\end{table}

\paragraph*{Trace-guided exploration} Figure~\ref{fig:tracecompare} demonstrates the impact of the trace guiding, in comparison to unguided, purely random search. With trace guiding, most agents follow a few distinct paths to discover the surrounding token clusters. Without guiding, the random-walk process ends up being equivalent to the nearest neighbor search: the likelihood of a token being discovered decreases as a square of distance from the origin, as the agents become more spread-out. The two marked regions A and B in Figure~\ref{fig:tracecompare}-right, illustrate this contrast: from the random walk density we see that region A is more thoroughly explored than B in spite of both having a similar Euclidean distance from the source. This translates to A being closer within the paradigm of optimal transport.

\paragraph*{Similarity ranking} Using word embeddings, we can evaluate the relations between words geometrically. For word similarity, the two widely used similarity metrics \cite{ dai2016unlocking, Berger2016cite2vec, Park2018conceptbuilding, ji2019visual} are the standard Euclidean distance $d_{Euclid}(v1, v2) = ||v1 - v2||$ and cosine similarity $d_{cos}(v1, v2) = v1 \cdot v2$. Cosine similarity assumes that two words represent directions on an N-dimensional hypersphere: the closer the directions, the more similar the words. The implication of this metric is that the spatial distance between two data points matters less than their direction from the origin.

To provide a \textbf{structure-aware measure}, we define our similarity metric by how reachable one data point is from another. In contrast with cosine and Euclidean similarity, MCPM similarity is defined by connectedness rather than their pure distance in space. In other words, the closeness of two data points is measured by whether other data points lay down a clear path between them. To this end we deploy agents from a chosen source point (Figure~\ref{fig:tracecompare}). If a point is found sufficiently close to an agent at any simulation step, a counter for that point is incremented. The resulting similarity to the source data point is proportional to the value of this counter at the end of the search (normalized over all discovered points).

To explore this new similarity measure, we choose $wind\_NOUN$ in W2V-300k to generate the similarity rankings based on the three different metrics. \new{We discuss how the metrics fare in section~\ref{sec:discussion}, but some distinctions can be gleaned from the word clouds of the top 30 discovered tokens shown in Figure~\ref{fig:wordcloud} and the difference list between MCPM ranking and other two rankings shown in Table \ref{tb:diff}.} The entries are ordered by descending difference between the MCPM ranking and Euclidean ranking.  

\begin{figure}[thb]
\begin{center}
\includegraphics[width=\columnwidth]{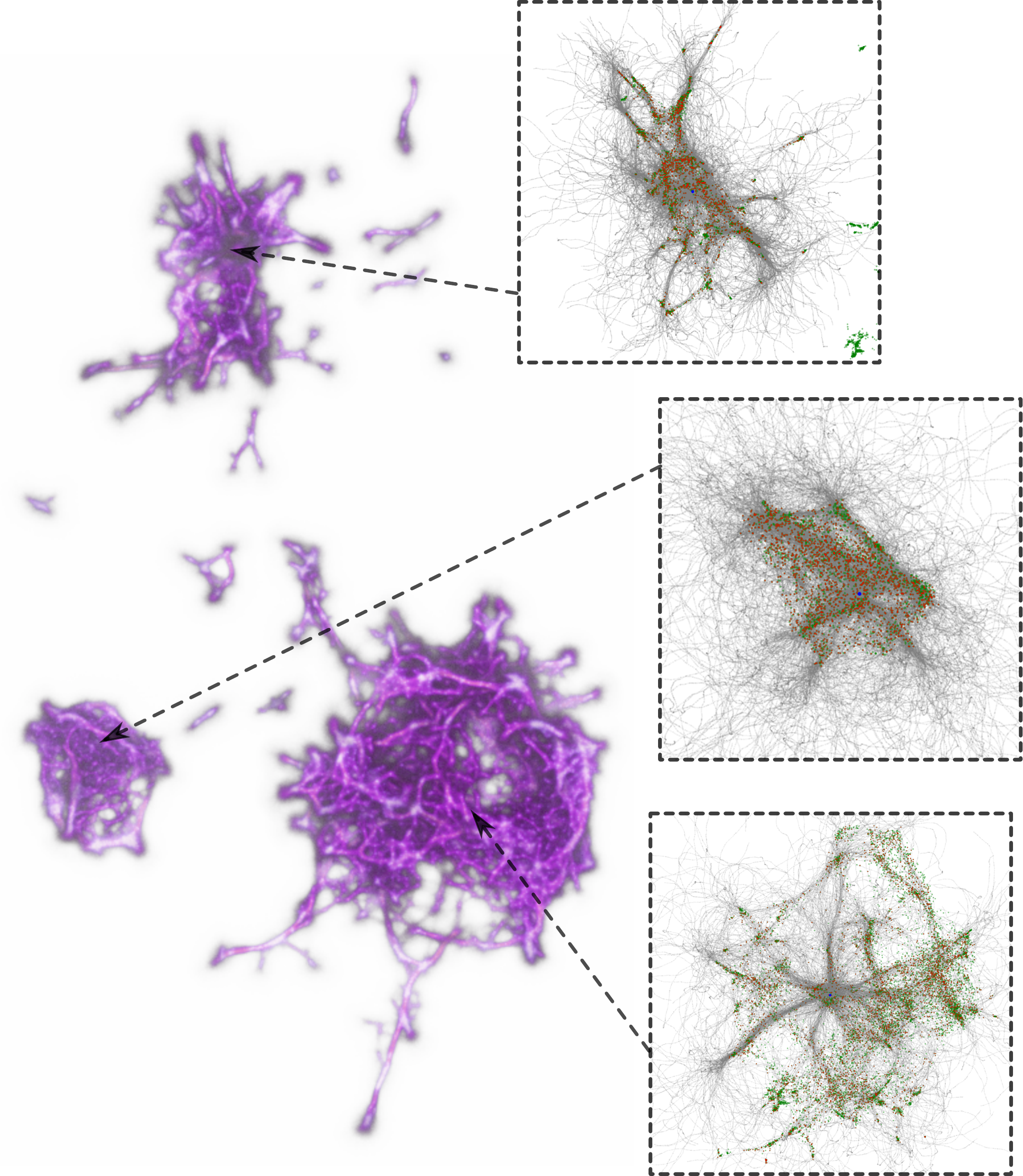}
\vspace{-6mm}
\caption{Visualization of intra-cluster exploration for BERT-back, starting in locations inside each respective cluster. We observe distinct topologies within each cluster, corresponding to the different contexts of word \textit{back} captured by the embedding (see Table \ref{tb:cluster}).}
\label{fig:localandexplore}
\vspace{-6mm}
\end{center}
\end{figure}

\begin{figure}[thb]
\begin{center}
\includegraphics[width=\columnwidth]{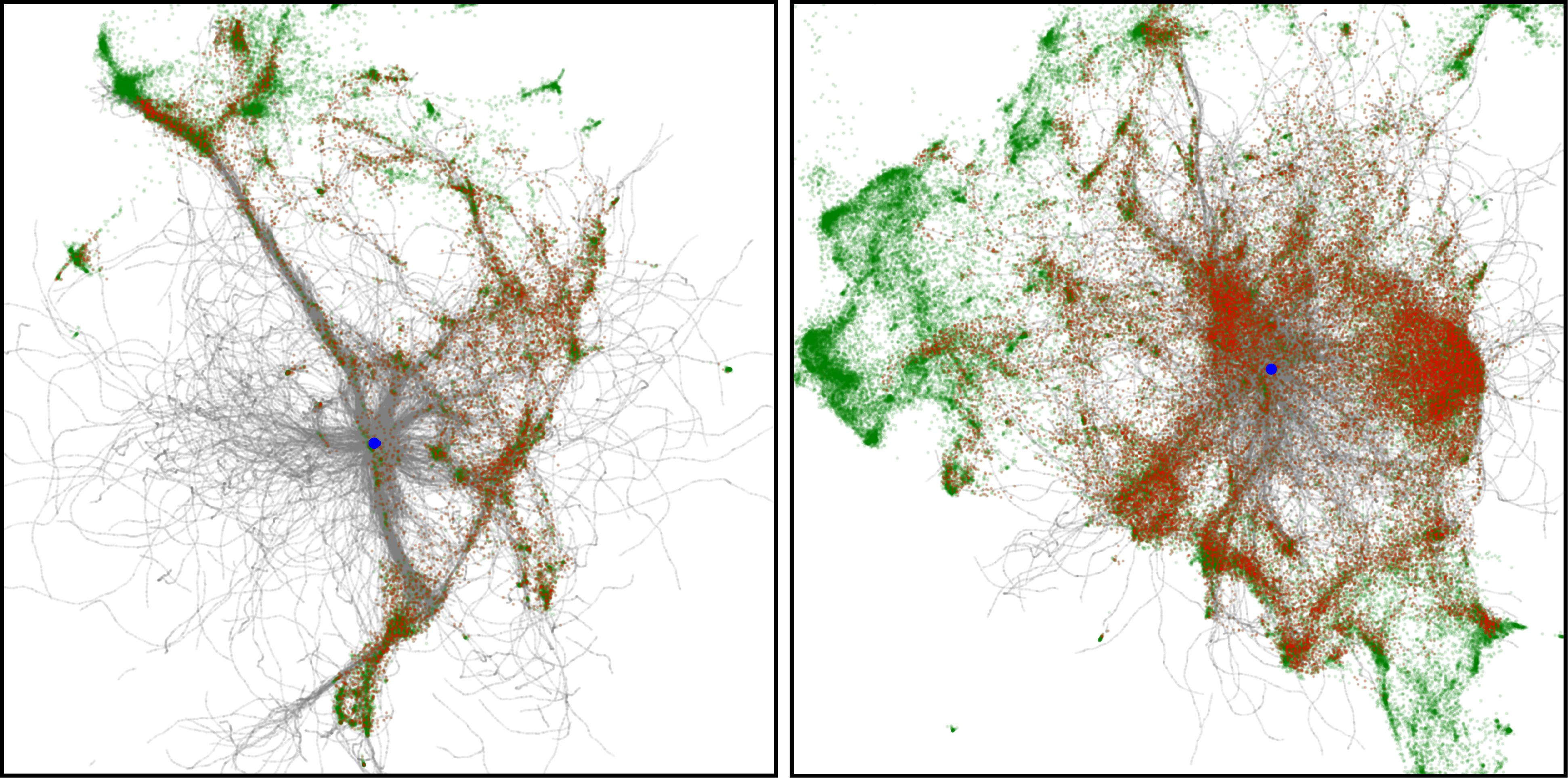}
\vspace{-3mm}
\caption{MCPM agent exploration results for \textit{class\_NOUN} (left) and \textit{research\_NOUN} (right). The comparison illustrates the difference between filament words (left) and knot words (right), which can be differentiated by the clearly articulated paths of the probe agents, as seen in the left figure.}
\label{fig:knotfilament}
\vspace{-6mm}
\end{center}
\end{figure}


\section{Discussion}\label{sec:discussion}

\paragraph*{Clusters in contextual embeddings} The contextual embeddings visualized in Figure~\ref{fig:local_embedding_3D} show a clear separation of clusters. MCPM acts as a robust clustering method here, in spite of their highly irregular shape. \new{We identify these clusters visually as components (sub-networks) interconnected by MCPM. Specifically, two tokens belong to the same cluster if one can be reached from the other by following the MCPM trace network.} To  explore the contents of these clusters, we sample several sample locations inside the embedding BERT-back, and then visualize the resulting searches in Figure~\ref{fig:localandexplore}.

The samples found within each cluster demonstrate clear differences in the word usage patterns (see Table~\ref{tb:cluster}). The irregular top cluster usages of \textit{back} as an indication of spatial relation. Both bottom-left and bottom-right clusters demonstrate \textit{back} as verbal particles used in phrasal verbs. The smaller cluster in the bottom-left shows usages of \textit{back} as a movement in time. Finally, the large bottom-right cluster indicates directionality of communication.

The separation of clusters as seen for polysemous words like \textit{back} and \textit{wind} (see Figure \ref{fig:local_embedding_3D}) indicates clear boundaries of these volumes and hints on the number of distinct contexts in which these words occur. MCPM similarity is useful here to not only identify the clusters, but to allow for their efficient exploration starting at arbitrary seed points within the clusters.

\paragraph*{Knot words versus filament words} With the structural information carried by the MCPM trace, we observe two distinct types of data points: \textbf{knot words} and \textbf{filament words}. As their names suggest, knot words are words positioned inside clusters and their emitted agents travel with no clear directionality, while filament words are positioned on the paths connecting clusters. For instance, the position of \textit{research\_NOUN}, visualized in Figure~\ref{fig:knotfilament}, stands in contrast with that of \textit{class\_NOUN}. We observe a clear difference in the distribution of agents’ travel directions.

\new{So far, the identification of these concepts has been based on the visual analysis (Section~\ref{sec:structure}). To enable automatic extraction of these properties, we propose to measure the directional statistics of agents spawned by a given query token. Per intuition provided by the visual analysis, the directional distribution of probe agents from a filament word should be bi-modal, while knot words should yield more complex multi-modal distributions. Based on these criteria, we plan to study these properties in bulk, and determine their origins and semantic significance.}

\paragraph*{Similarity ranking} The three compared rankings emphasize different mathematical relations. Cosine similarity measures orientation with respect to origin, Euclidean similarity measures geodesic distance in a homogeneous space, and MCPM similarity builds on the optimal-transport throughput. Our aim is to explore how these properties can be used for a more intuitive way to understand machine-generated language representations. Based on the similarity rankings, different properties seem to highlight some word tokens that others do not. 
Curiously, the cosine similarity shows many items not found in other rankings. For the query \textit{wind}, words such as \textit{budget}, \textit{ornamentation} and \textit{overpay} seem rather out-of-place but are still placed highly in the ranks. Euclidean and MCPM rankings have much more agreeable candidates in higher ranks such as \textit{gust}, \textit{hurricane-force} and \textit{anemometer}. Many of them are specialized terms in climatology. \new{This implies} that geodesic distance is an important factor when considering word closeness in W2V-300k.

We also observe a divergence between Euclidean ranking and MCPM ranking in Table~\ref{tb:diff}. MCPM similarity manages to capture words quite far away from the source point. Interestingly, the two words \textit{unseasonable} and \textit{near-record} are still consistent with the general climatological theme of the similarity ranking. Words like \textit{anticyclone} and \textit{intertropical} are very similar to many of the top candidates such as \textit{cyclogenesis} and \textit{non-tropical}. This finding seems to suggest that spatial distance is also an imperfect measurement of similarity, and the measurement of similarity should also consider the throughput between words. This of course needs to be verified by a broader, quantitative study in the future.

It is also important to realize that the definition of similarity as such is rather vague semantically. Computational linguistics distinguishes between the concept of association and similarity. While one would agree that \textit{tropical} is more similar to \textit{wind} than \textit{overpay}, we can only claim they are more similar because it’s easier to associate the word \textit{tropical} to \textit{wind}. At the same time, the word \textit{gust} and \textit{squall} can be said to be associated with, but also similar to \textit{wind} \cite{hill2015simlex}. It remains \new{an open question} whether there is a way to extract the distinction between associated and similar words in word embeddings.

\new{Since the Monte Carlo exploration is a stochastic process by definition, it is important to address the stability of our similarity ranking. On the word embedding level, studies have shown embedding algorithms to be non-deterministic even with the same training data and configuration~\cite{heimerl2018interactive}. On the trace generation level, and the subsequent trace-guided exploration by probe agents, we solely rely on converged aggregate solutions. In the MCPM similarity ranking, we observe that the results are fairly stable as each ranking only shifts by 1 to 2 places between different results. From this we conclude that the probabilistic nature of our framework does not render the results unreliable, especially considering that even human-produced rankings tend to have significant inter-subject variation.}

\paragraph*{Origin of the structures} Ultimately, an important question to consider is to what extent are the discovered transport networks and resulting structures inherent to the embeddings. \new{It has been shown that non-linear methods such as t-SNE and UMAP distort pairwise relationship between embeddings, while PCA can introduce false positive parallel pairs in its result\cite{8019864}.} The important consideration is therefore the choice of the projection method: we chose UMAP because of its recognized ability to preserve both global and local structures, as well as the number of components in the source data. This is in contrast to other methods like t-SNE and PCA, which are known to destroy all global and local structure, respectively. Our experiments with different configurations of UMAP also showed that regardless of the resulting projection shape, the structures were present and reliably recovered by MCPM. Finally, the fact that MCPM similarity yields reasonable results even in 3D is very encouraging, and motivates us to look for solutions that operate in the native, high-dimensional embedding space.


\section{Conclusion} \label{sec:future}

In this exploratory paper, we have investigated how a bio-inspired random walk model pioneered for structure-finding in astronomy \cite{burchett2020revealing, elek2020monte} can help us identify and visualize interesting geometric configurations in word embeddings. We show that this method can reveal a range of potential structural factors in the embedding space, including the number of components for a word, as well as its neighborhood structure. Our approach combines a holistic view of an embedding dataset as an optimal transport network, while enabling us to pay attention to structures on the word and cluster level.

We believe these results are a promising start to a new line of research, and a great deal of work lies ahead of us to understand if these structures correspond to meaningful semantic distinctions. Building a robust interpretation for the information encoded in the embeddings would also improve natural language processing tasks like word similarity and disambiguation.

We are currently in the process of redesigning the MCPM simulation to run in the original high-dimensional spaces instead of a reduced one, to overcome the question which of the detected structures are valid and which ones are induced by the dimensionality reduction. This challenging extension is also attractive for other application domains which make use of embeddings, for instance recommender engines, music processing, game state space exploration, and generative systems.

\new{We also plan to more deeply probe the filament\,/\,knot distinction. We intend to develop a concrete mathematical characterization of these two concepts -- something that is needed to automatically extract them and explore to what extent is this information semantically meaningful and, in the long run, to understand where it comes from. This understanding could be instrumental for structural comparison of different text corpora and even entire languages.}

\bibliographystyle{abbrv-doi}

\bibliography{bibliography}
\end{document}